\documentclass[10pt,twocolumn,letterpaper]{article}

\usepackage{cvpr}

\definecolor{dark_blue}{rgb}{0, 0, 0.7}

\definecolor{dark_green}{rgb}{0, 0.5, 0}

\newcommand{\paragrapht}[1]{\vspace{-10pt}\paragraph{#1}}
\newcommand{\hlrow}{\rowcolor{black!6}}
\newcommand{\ours}{Chrono\xspace}

\definecolor{cvprblue}{rgb}{0.21,0.49,0.74}
\definecolor{changecolor1}{RGB}{255,0,0}

\usepackage[pagebackref,breaklinks,colorlinks,allcolors=cvprblue]{hyperref}
\usepackage{multicol}
\usepackage{multirow}
\usepackage{colortbl}
\usepackage{lipsum}

\def\paperID{14549} %
\def\confName{CVPR}
\def\confYear{2025}

\title{Exploring Temporally-Aware Features for Point Tracking}

\author{Inès Hyeonsu Kim\textsuperscript{1*}\qquad
Seokju Cho\textsuperscript{1*}\qquad
Jiahui Huang\textsuperscript{2}\qquad
Jung Yi\textsuperscript{1}\qquad\\
Joon-Young Lee\textsuperscript{2}\qquad
Seungryong Kim\textsuperscript{1}\qquad\\[10pt]
\textsuperscript{1}KAIST AI\qquad \textsuperscript{2}Adobe Research
}

\begin{document}
\maketitle
\begin{abstract}
Point tracking in videos is a fundamental task with applications in robotics, video editing, and more. While many vision tasks benefit from pre-trained feature backbones to improve generalizability, point tracking has primarily relied on simpler backbones trained from scratch on synthetic data, which may limit robustness in real-world scenarios. Additionally, point tracking requires temporal awareness to ensure coherence across frames, but using temporally-aware features is still underexplored. Most current methods often employ a two-stage process: an initial coarse prediction followed by a refinement stage to inject temporal information and correct errors from the coarse stage. These approach, however, is computationally expensive and potentially redundant if the feature backbone itself captures sufficient temporal information.

In this work, we introduce \textbf{\ours}, a feature backbone specifically designed for point tracking with built-in temporal awareness. Leveraging pre-trained representations from self-supervised learner DINOv2 and enhanced with a temporal adapter, \ours effectively captures long-term temporal context, enabling precise prediction even without the refinement stage. Experimental results demonstrate that \ours achieves state-of-the-art performance in a refiner-free setting on the TAP-Vid-DAVIS and TAP-Vid-Kinetics datasets, among common feature backbones used in point tracking as well as DINOv2, with exceptional efficiency. Project page: \url{https://cvlab-kaist.github.io/Chrono/} 
\end{abstract}
\let\thefootnote\relax\footnotetext{$^*$Equal contribution}

\begin{figure}[t]
    \centering
    \includegraphics[width=\linewidth]{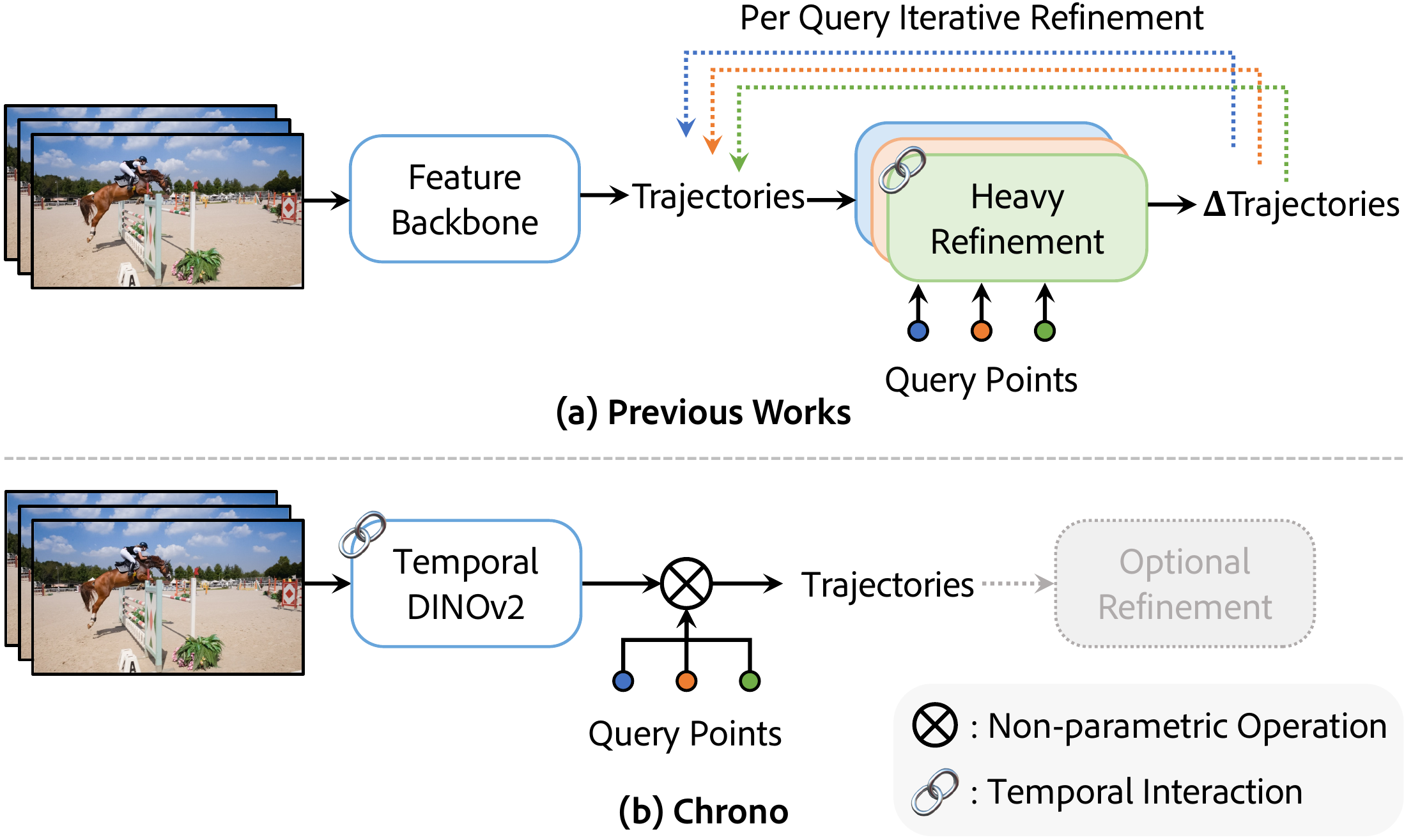}
    \caption{\textbf{\ours} is a highly precise, temporally-aware feature backbone specifically designed for point tracking. Unlike prior methods that rely on heavy iterative refinement for each query point~\cite{harley2022particle, doersch2022tap, doersch2023tapir, cho2024local, karaev2023cotracker} or test-time optimization~\cite{tumanyan2025dino, wang2023tracking}, \ours achieves competitive tracking performance through simple feature matching using a non-parametric operation. Moreover, its performance can be further improved by an optional refinement step.
    }
    \label{fig:teaser}
\vspace{-10pt}
\end{figure}

\section{Introduction}
\label{sec:intro}
Point tracking aims to track any point within a casual video, which has wide-ranging applications such as robotics~\cite{vecerik2023robotap}, video editing~\cite{huang2023inve}, and view synthesis~\cite{wang2024shape}. It involves establishing correspondences of specific points across frames to determine their positions over time and whether they are visible or occluded. Accurate point tracking is challenging due to complex motions, occlusions, and deformations~\cite{doersch2023tapir, cho2024local, doersch2022tap, doersch2024bootstap}. Achieving robust tracking requires effective matching of points across frames and a comprehensive understanding of temporal dynamics.

\begin{figure}[t]
    \centering
    \includegraphics[width=\linewidth]{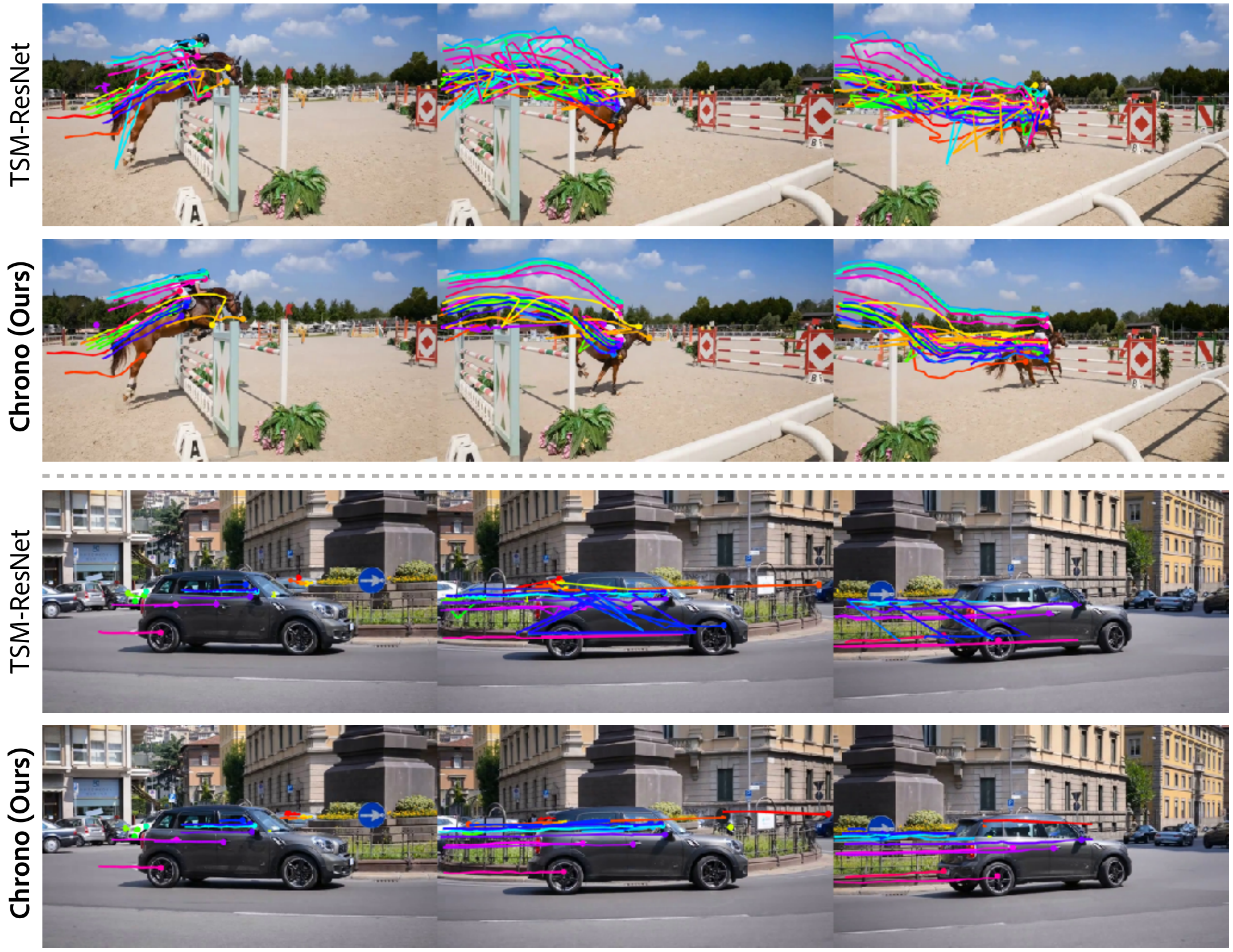}
    \caption{\textbf{Comparison of initial trajectories from Chrono and existing feature backbone for point tracking.} \ours demonstrates a significant improvement over existing feature backbones such as TSM-ResNet~\cite{lin2019tsm}.}
    \label{fig:teaser-qual}
\vspace{-10pt}
\end{figure}

In many computer vision tasks such as segmentation, object detection, and depth estimation, it is often beneficial to use pre-trained feature backbones along with task-specific heads~\cite{Cheng_2022_CVPR, carion2020end, bhat2023zoedepth, he2017mask, chen2017deeplab}. These backbones, trained on large-scale datasets, offer robust representations that enhance generalizability to real-world data. In contrast, point tracking models often rely on simpler backbones such as ResNet~\cite{he2016deep} and are frequently trained from scratch~\cite{doersch2022tap, doersch2023tapir, cho2024local, harley2022particle, karaev2023cotracker} on synthetic dataset~\cite{greff2022kubric, zheng2023pointodyssey}.  This difference raises an important question: \textit{Could point tracking similarly benefit from adopting pre-trained feature backbones?}

While these backbones are often pre-trained on single-image tasks~\cite{oquab2023dinov2, tumanyan2025dino, liu2021swin, dosovitskiy2020image, zhai2022scaling}, point tracking inherently demands temporal awareness to maintain coherence across video frames. Unlike single-image tasks, point tracking requires consistent point matching over time, necessitating a model capable of capturing temporal information~\cite{cho2024local, karaev2023cotracker, karaev2024cotracker3, li2024taptr, vecerik2023robotap, doersch2024bootstap}. Therefore, features used in point tracking also need to be temporally aware.
 
Despite its critical role, temporally-aware backbone has been relatively under-explored in this area. Early efforts, such as simply incorporating TSM-ResNet~\cite{lin2019tsm} in point tracking models~\cite{doersch2022tap,doersch2023tapir}, yielded limited improvements. This outcome may be attributed to two factors: First, training on small synthetic datasets from scratch~\cite{greff2022kubric} may limit generalization to real-world data. Second, TSM-ResNet’s temporal awareness, constrained to adjacent frames, may not provide long-term temporal context for point tracking.

On the other hand, point tracking methods often adopt a two-stage process~\cite{doersch2023tapir,cho2024local,doersch2024bootstap}: an initial stage that predicts coarse tracks directly from the simple feature backbone such as ResNet, followed by a refinement stage that introduces temporal information by analyzing trajectories across frames and iteratively refining the prediction. This pipeline has become standard, with refinement methods aiming to mitigate noise and enhance track smoothness~\cite{cho2024local, karaev2023cotracker, karaev2024cotracker3, li2024taptr, vecerik2023robotap, doersch2024bootstap}. However, these refinement steps can be computationally demanding and query-dependent~\cite{cho2024local, cho2024flowtrack}, often requiring temporal refinement for each query point individually, which impacts efficiency. Moreover, a strong reliance on post-hoc refinement adds a substantial burden in correcting initial prediction errors from the first stage. This error could potentially be mitigated by leveraging features with built-in temporal awareness.

To overcome these limitations, we introduce a temporally-aware feature backbone for point tracking, called \textbf{\ours}, as illustrated in Figure~\ref{fig:teaser}. Our key insights are twofold: First, we leverage robust feature representations learned from large-scale, real-world datasets. We employ DINOv2~\cite{oquab2023dinov2}, a pre-trained model known for strong feature representations across various tasks such as segmentation, classification, and localization~\cite{leroy2024grounding, oquab2023dinov2}. However, these representations are not directly compatible with point tracking due to a lack of temporal awareness. Second, we incorporate temporal awareness by designing a temporal adapter that enables the pre-trained backbone to process data with temporal context, without losing its learned knowledge. This enables the use of these strong feature backbones for point tracking tasks. Unlike the adjacent-frame context used in TSM-ResNet~\cite{lin2019tsm}, our approach incorporates a temporal context that is six times longer, enabling the capture of complex dynamics and enhancing tracking performance, as shown in Figure~\ref{fig:teaser-qual}.

Experimental results show that tracks estimated from \ours features, computed solely using soft argmax without any learnable layers or additional temporal information after feature extraction, significantly outperform traditional feature backbones commonly used in point tracking as well as DINOv2~\cite{oquab2023dinov2}. Specifically, \ours achieves a +20.6\%p increase in the position accuracy on the TAP-Vid-DAVIS~\cite{doersch2022tap} dataset compared to TSM-ResNet-18~\cite{lin2019tsm}. Additionally, our method demonstrates superior efficiency and comparable performance compared to point tracking models equipped with refiners that inject temporal information post-feature extraction. Despite the absence of learnable layers after feature extraction, \ours achieves high precision, underscoring that directly embedding temporal information within the features is both efficient and powerful. Optionally, we can add iterative refinement on top of our backbone, which can further enhance the performance of the refiner compared to when an existing backbone is used.

In summary, our contributions are as follows:
\begin{itemize}
    \item We highlight the lack of temporally-aware feature backbones and the reliance on computationally intensive refinement processes in current point tracking methods.
    \item We propose a feature backbone designed for point tracking that incorporates a long-range temporal adapter, enhancing temporal awareness over extended sequences.
    \item We demonstrate that our backbone produces accurate initial tracks in simple and effective manner, reducing the need for extensive refinement and achieving both improved performance and efficiency.
\end{itemize}

\section{Related Work}
\label{sec:related_works}

\paragraph{Point tracking.} 
PIPs~\cite{harley2022particle} independently tracks a point by fetching a local correlation around the point estimate and gradually refining the tracking result through iteration. TAP-Net~\cite{doersch2022tap} utilizes a shallow TSM-ResNet~\cite{lin2019tsm} for point tracking, with lightweight layers added to the backbone. TAPIR~\cite{doersch2023tapir} integrates TAP-Net with the iterative refinement from PIPs, modifying the architecture to a convolution-based model for temporal processing. CoTracker tracks multiple points simultaneously, modeling their dependencies with a Transformer architecture. LocoTrack~\cite{cho2024local} achieves improved correspondence using local 4D correlation, inspired by dense matching literature~\cite{cho2021cats}. TAPTR~\cite{li2024taptr} introduces the DETR-like~\cite{carion2020end} architecture for point tracking. Another line of work employs test-time optimization with regularization, such as tracking smoothness, geometric constraints, and cycle consistency~\cite{wang2023tracking, tumanyan2025dino}. While these methods focus on designing better architectures for track refinement, the exploration of more effective feature backbones has been relatively underexplored. Our work focuses on developing an improved feature backbone.

\paragrapht{Feature backbone in point tracking.}
Early works on point tracking~\cite{doersch2022tap} explored the performance of self-supervised feature backbones, such as VFS~\cite{xu2021rethinking}, as well as temporally-aware features like TSM-ResNet~\cite{lin2019tsm}. However, the exploration of feature backbones has been relatively underdeveloped compared to iterative track refinement~\cite{doersch2023tapir,karaev2023cotracker,cho2024local,cho2024flowtrack}. DINO-Tracker~\cite{tumanyan2025dino} explored the use of DINO in point tracking but it requires an hour of optimization for each video. Recently, \cite{aydemir2024can} investigated the tracking capabilities of foundational models, such as Stable Diffusion~\cite{rombach2022high} and DINOv2~\cite{oquab2023dinov2}, though the potential to extend these models for temporal applications remains unexplored.

\paragrapht{Adapting large feature backbone.}
With recent advances in large-scale self-supervised training~\cite{he2022masked, oquab2023dinov2}, effectively transferring its vast knowledge has become a key challenge. Specifically, the strong features of DINOv2 have been adapted to various tasks. \cite{salehi2023time} fine-tunes DINO for video tasks in a self-supervised manner. Additionally, DINO has been applied to various correspondence tasks~\cite{amir2021deep, mariotti2024improving,shtedritski2023learning, leroy2024grounding,gupta2023asic} and has demonstrated robustness in establishing correspondence and semantic segmentation~\cite{hamilton2022unsupervised, amir2021deep, melas2022deep}, showing its rich semantics. We focus on adapting this representation to the point tracking task, enhancing its temporal consistency while preserving its pre-trained knowledge.

\section{Method}
\begin{figure}
    \centering
    \includegraphics[width=\linewidth]{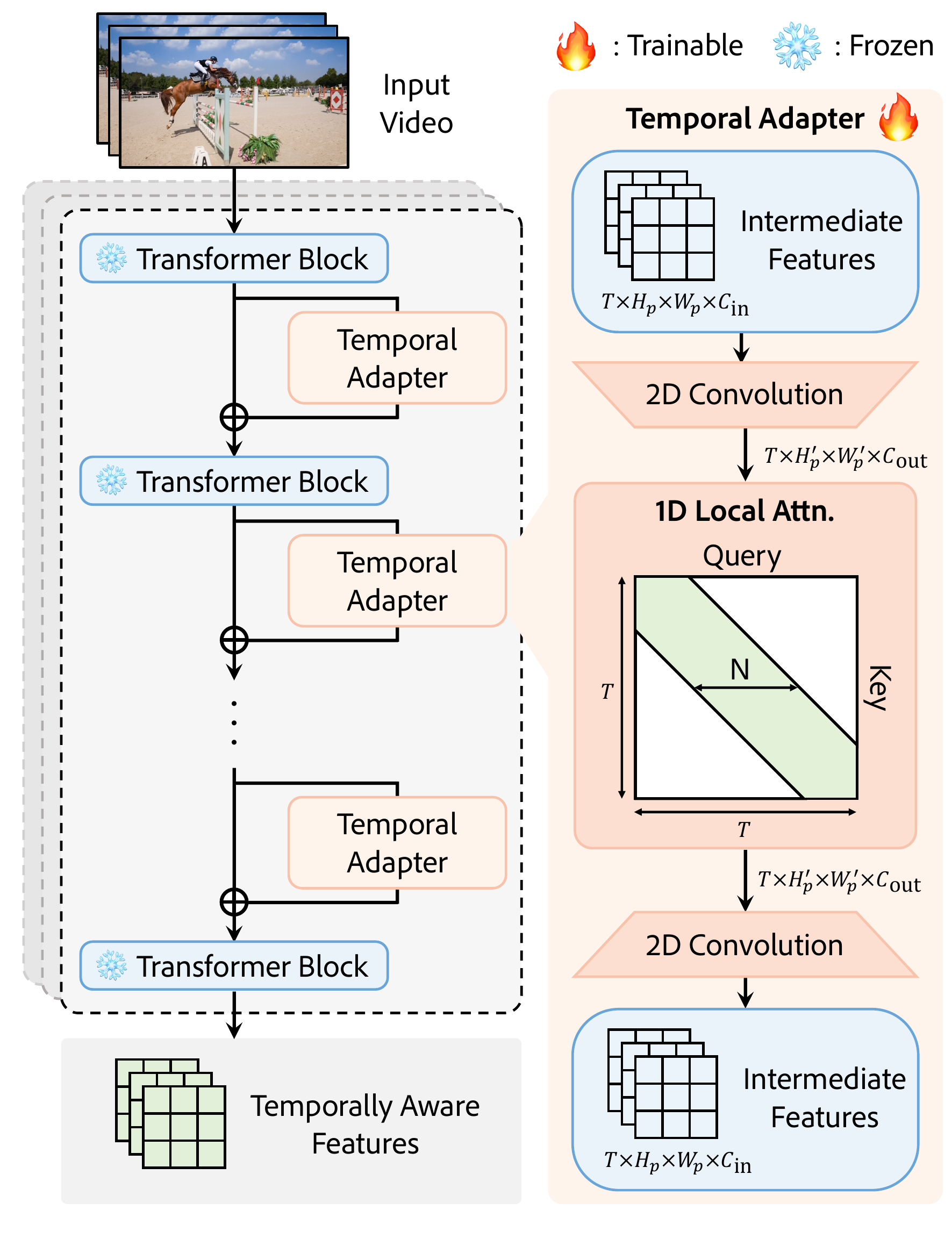}
    \vspace{-15pt}
    \caption{\textbf{Overall architecture of \ours.} Temporal adapters between transformer blocks use 2D convolution and 1D local attention along the temporal axis to output temporally-aware features.}
    \label{fig:main_arch}
    \vspace{-15pt}
\end{figure}

In this section, we introduce our temporally aware feature backbone. We begin by formally defining the point tracking problem and its associated challenges. We then describe the design of our temporal adapter and its integration with a pre-trained backbone to incorporate temporal awareness. Finally, we present our approach for adapting the pre-trained backbone and training the temporal adapter for point tracking. 

\subsection{Preliminaries and Motivation}

\paragraph{Task definition.} Point tracking~\cite{li2024taptr, doersch2023tapir, doersch2024bootstap, doersch2022tap, karaev2023cotracker, cho2024local, cho2024flowtrack} in videos involves establishing correspondences of specific points across frames, determining their positions over time, and identifying whether they are visible or occluded. Formally, given a video sequence $\{ \mathcal{I}_t \}_{t=0}^{T-1}$, where $\mathcal{I}_t \in \mathbb{R}^{H \times W \times 3}$ represents the $t$-th frame of height $H$ and width $W$, and a query point $q = (x_q, y_q, t_q) \in \mathbb{R}^3$ at frame $t_q$, the goal is to produce a trajectory $\mathcal{T} = \{ \hat{p}_t \}_{t=0}^{T-1}$, where $\hat{p}_t \in \mathbb{R}^2$ denotes the estimated position of the point at time $t$, and an associated visibility probability $\mathcal{V} = \{ v_t \}_{t=0}^{T-1}$, where $v_t \in [0,1]$ indicates the likelihood of the point being visible at frame $t$. In this work, we focus on predicting accurate positions rather than occlusions, as our primary goal is to achieve better point correspondences.

\paragrapht{Motivation.} 
Accurate point tracking in videos requires robust point matching across frames, a challenging task due to complex motions, occlusions, deformations, and scale variations~\cite{wang2023tracking, karaev2023cotracker, cho2024local}. Because videos consist of sequences over time, considering both spatial and temporal aspects is essential to understand point dynamics. More specifically, a good feature for point tracking must satisfy two key criteria. Spatially, it should effectively model complex real-world data with powerful feature representations vital for robust matching, the cornerstone of point tracking~\cite{cho2024local}. Temporally, it must capture complex motions and understand the dynamics of points over time to handle their movement across frames.

To fulfill the spatial criterion, we utilize DINOv2~\cite{oquab2023dinov2}, a self-supervised model trained on large-scale real-world data, renowned for its robust feature representations across various tasks~\cite{oquab2023dinov2, bochkovskii2024depth, leroy2024grounding}. However, DINOv2 lacks inherent temporal awareness, which is essential for point tracking in videos. To address this limitation, we introduce a temporal adapter that supplements DINOv2's features with temporal information. This adapter integrates temporal understanding directly into the feature extraction process, creating a feature backbone tailored for point tracking that combines strong spatial representations with temporal dynamics.

\subsection{Temporal Adapter}
Temporal awareness is essential for accurately tracking points across frames, as it enables the model to capture motion patterns and temporal dependencies. To embed this capability within DINOv2, we design a temporal adapter that allows the backbone to incorporate information from adjacent frames, enhancing its temporal sensitivity. Figure~\ref{fig:main_arch} illustrates the overall architecture of \ours. 

\paragrapht{Design of the temporal adapter.} Our temporal adapter is placed between each transformer block of DINOv2~\cite{oquab2023dinov2} to enhance temporal modeling across multiple feature levels. By computing adjacent features from other frames at once, the adapter connects features from different time steps, enabling it to recognize motion across frames.

Each adapter follows a bottleneck structure inspired by the ResNet~\cite{he2016deep} architecture. It begins with a 2D convolutional layer \(\text{Conv2D}^{\text{down}}\) with stride \( s \) that reduces the spatial dimensions from \( H_p \times W_p \times C_{\text{in}} \) to a compact \( H_p' \times W_p' \times C_{\text{out}} \), where \( H_p' = {H_p}/{s} \) and \( W_p' = {W_p}/{s} \). This spatial downsampling step optimizes the representation for efficient processing while expanding the spatial receptive field for subsequent operations. This is formally defined as:
\begin{equation}
\mathbf{f}^{\text{down}}_{t} = \text{Conv2D}^{\text{down}}(\mathbf{f}^{\text{in}}_{t}),
\end{equation}
where \( \mathbf{f}^{\text{in}}_{t} \in \mathbb{R}^{H_p \times W_p \times C_{\text{in}}} \) and \( \mathbf{f}^{\text{out}}_{t} \in \mathbb{R}^{H_p' \times W_p' \times C_{\text{out}}} \) are the input and output feature maps of \(\text{Conv2D}^{\text{down}}\) at time \( t \), respectively.

After downsampling, we apply a temporal attention layer that captures dependencies within a local temporal window of size \( N \), focusing on temporal correlations across time~\cite{roy2021efficient, beltagy2020longformer}. The local window attention is carefully calibrated to balance computational efficiency with effective motion capture, ensuring a broad enough temporal window without excessive resource demands. This layer operates over the neighboring frames within the window \( [t - k, t + k] \), where \( k = \frac{N - 1}{2} \). For each spatial location \( (x, y) \), we compute attention weights based on the similarity between the query vector at time \( t \) and key vectors from neighboring times. The aggregated feature at time \( t \) and location \( (x, y) \), referred to as \(\mathbf{f}^{\text{attn}}_{t}(x, y)\), is computed as a weighted sum of the value vectors from these neighboring frames:
\begin{equation}
\mathbf{f}^{\text{attn}}_{t}(x, y) = \sum_{n = -k}^{k} \alpha^{(t,n)}(x, y) \cdot \mathbf{V}_{t+n}(x, y),
\end{equation}
where \( \alpha^{(t,n)}(x, y) \) represents the attention weights for time offset \( n \), computed as:
\begin{equation}
\alpha^{(t,n)}(x, y) = \frac{\exp\left( \mathbf{Q}_{t}(x, y) \cdot \mathbf{K}_{t+n}(x, y) / \sqrt{d_k} \right)}{\sum_{n' = -k}^{k} \exp\left( \mathbf{Q}_{t}(x, y) \cdot \mathbf{K}_{t+n'}(x, y) / \sqrt{d_k} \right)},
\end{equation}
where \(d_k\) is a scaling factor~\cite{vaswani2017attention}, $\mathbf{Q}$, $\mathbf{K}$, and $\mathbf{V}$ are projections of \(\mathbf{f}^{\text{down}}\), obtained through the linear projection layers \( \mathbf{W}_Q \), \( \mathbf{W}_K \), and \( \mathbf{W}_V \).

After temporal attention, another 2D convolutional layer restores the spatial dimensions back to \( H_p \times W_p \times C_{\text{in}} \):
   \begin{equation}
   \mathbf{f}^{\text{up}}_{t} = \text{Conv2D}^{\text{up}}(\mathbf{f}^{\text{attn}}_{t}),
   \end{equation}
where \( \mathbf{f}^{\text{up}}_{t} \in \mathbb{R}^{H_p \times W_p \times C_{\text{in}}} \). Finally, a residual connection adds the original input feature map to the output of the adapter to preserve the feature representation of DINOv2:
  \begin{equation}
   \mathbf{f}^{\text{out}}_{t} = \mathbf{f}^{\text{up}}_{t} + \mathbf{f}^{\text{in}}_{t}.
    \end{equation}

\subsection{Point Prediction with the Feature Backbone}
\label{method: point_prediction}
\begin{figure}
    \centering
    \includegraphics[width=\linewidth]{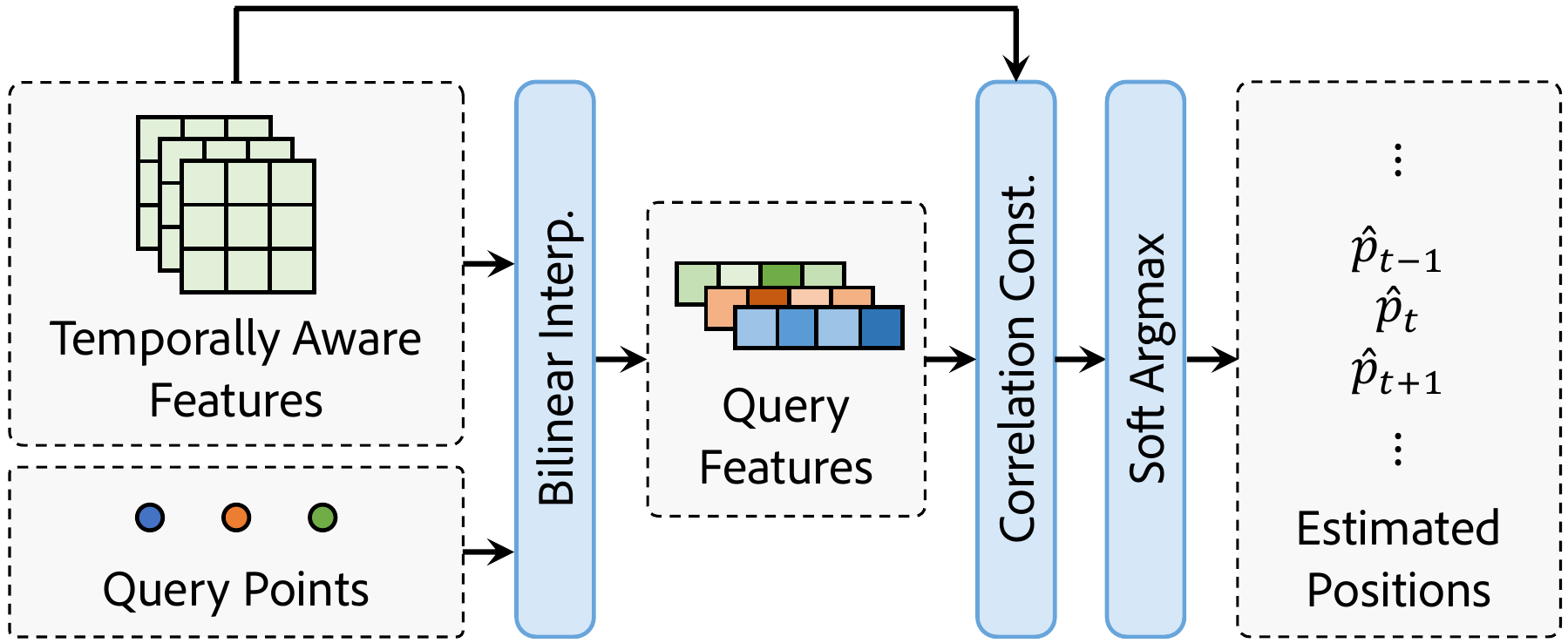}
    \caption{\textbf{Point track prediction.} To predict point positions, we simply match the query points with features from other frames, without using any learnable layers.
    }
    \label{fig:point_prediction}
    \vspace{-10pt}
\end{figure}

As shown in Figure~\ref{fig:point_prediction}, we track a query point by simply performing feature matching using features from \ours. This approach avoids the need for learnable modules such as iterative refinement, as used in~\cite{cho2024local,karaev2023cotracker,cho2024flowtrack,doersch2023tapir}.

\paragrapht{Correlation construction.}
Given the query point $q = (x_q, y_q, t_q)$, we extract the query feature $\mathbf{f}_q$ from the feature map of frame $t_q$ at position $(x_q, y_q)$ using bilinear interpolation. For each frame $t$, we compute the correlation map $\mathcal{C}_t$ by calculating the cosine similarity between $\mathbf{f}_q$ and the feature map $\mathbf{f}_t$ at every spatial location:
\begin{equation}
\mathcal{C}_t(x, y) = \frac{\mathbf{f}_q^\top \mathbf{f}_t(x, y)}{\|\mathbf{f}_q\| \|\mathbf{f}_t(x, y)\|},
\end{equation}
where $\mathbf{f}_t(x, y)$ is the feature vector at position $(x, y)$ in frame $t$, \( \mathbf{f}_q^\top \mathbf{f}_t(x, y) \) denotes the dot product between \( \mathbf{f}_q \) and \( \mathbf{f}_t(x, y) \), and \( \|\mathbf{f}_q\| \) and \( \|\mathbf{f}_t(x, y)\| \) indicate the Euclidean norms of \( \mathbf{f}_q \) and \( \mathbf{f}_t(x, y) \), respectively. This correlation map \( \mathcal{C}_t \) quantifies the similarity between the query feature and features across all spatial locations in each frame \( t \).

\paragrapht{Point prediction.}

To estimate the position of points in frame $t$, we apply the soft-argmax operation~\cite{lee2019sfnet} to the correlation map $C_t$. The soft-argmax computes a weighted average of all spatial positions, with weights given by the softmax of the correlation values:
\begin{equation}
\hat{\mathbf{p}}_t = \sum_{(x, y)} \sigma\left( C_t(x, y) \right) \cdot (x, y),
\end{equation}
where $\sigma\left( C_t(x, y) \right)$ is the softmax over all spatial positions within frame \(t\):
\begin{equation}
\sigma\left( C_t(x, y) \right) = \frac{\exp\left( C_t(x, y) \cdot \tau \right)}{\sum_{(x', y')} \exp\left( C_t(x', y')  \cdot \tau\right)},
\end{equation}
where \(\tau\) is a softmax temperature. The soft argmax is non-learnable and provides a differentiable method to estimate the point's position based on the correlation map. To enhance precision, we mask out positions more than $M$ pixels away from the maximum correlation value, focusing the soft argmax computation on a local neighborhood and reducing the influence of irrelevant regions~\cite{lee2019sfnet}.

\paragrapht{Training.}
To train \ours, we use the Huber loss~\cite{huber1992robust} to supervise the estimated positions. The Huber loss is chosen for its robustness to outliers compared to the squared error loss. At each time step \(t\), the loss is defined as:
\begin{equation}
\mathcal{L}_{\text{Huber}}(\hat{{p}}_t, {p}_t) = \begin{cases}
\frac{1}{2} \left\| \hat{{p}}_t - {p}_t \right\|^2, \! &\text{if } \left\| \hat{{p}}_t - {p}_t \right\| \leq \delta, \\
\delta \cdot \left( \left\| \hat{{p}}_t - {p}_t \right\| - \frac{1}{2} \delta \right), & \text{otherwise},
\end{cases}
\end{equation}
where ${p}_t$ is the ground truth position, and $\delta$ is a threshold parameter. For occluded points indicated by the ground truth visibility status $v_t$, we exclude the loss computation, effectively masking out those time steps:
\begin{equation}
\mathcal{L}_t = (1 - v_t) \cdot \mathcal{L}_{\text{Huber}}(\hat{{p}}_t, {p}_t).
\label{eq: huber}
\end{equation}
By minimizing this loss over all time steps and query points, we train the backbone to produce features for predicting accurate tracked positions.

\section{Experiments}
\begin{figure*}[h]
    \centering
    \includegraphics[width=\textwidth]{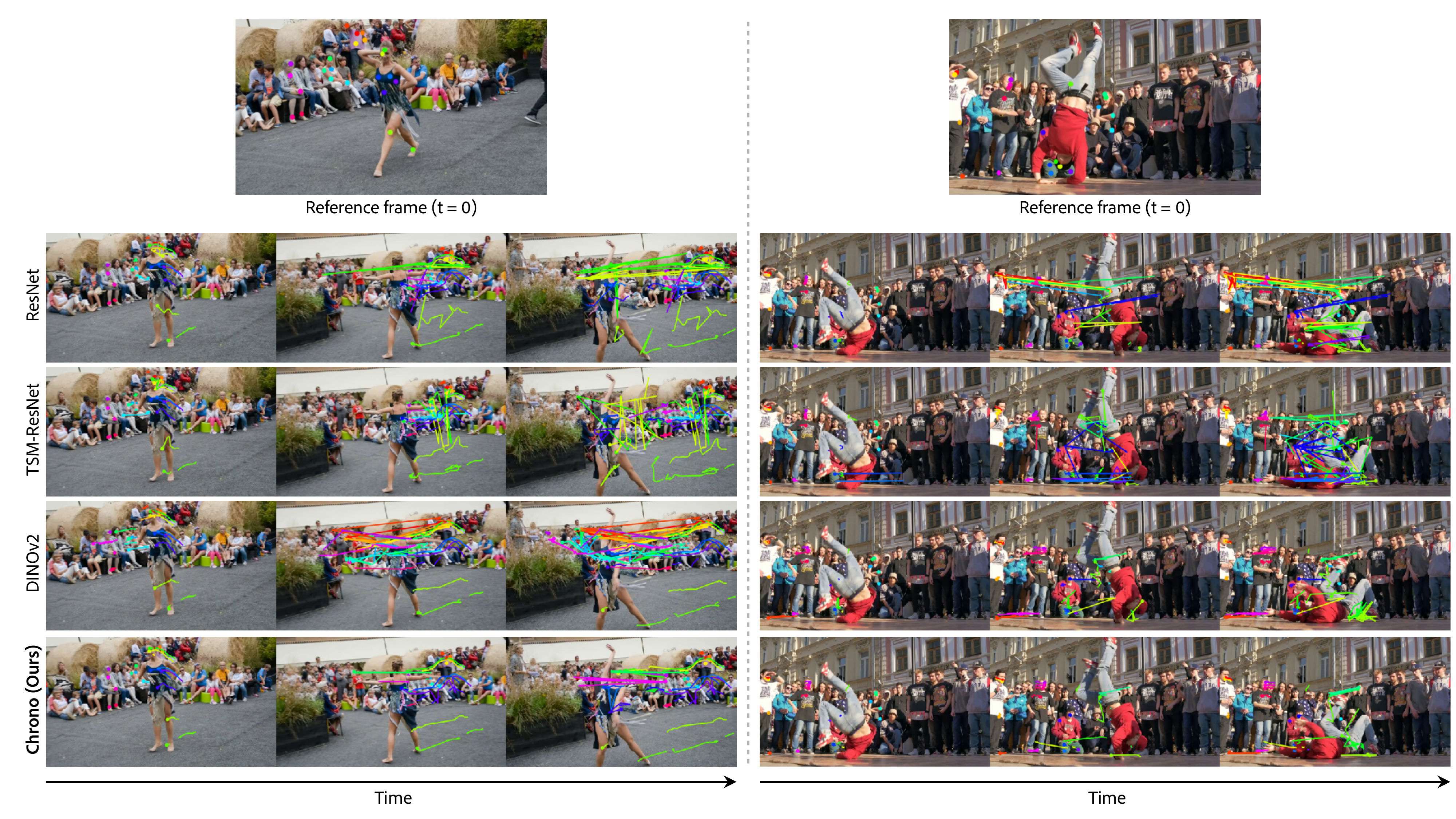}
    \vspace{-20pt}
    \caption{\textbf{Qualitative comparison of complex real-world video tracking.} We qualitatively compare the results generated by \ours with those from other commonly used backbones in point tracking and as well as DINOv2. Our model demonstrates better smoothness and precision than other competitors.}
    \label{fig:main_qual}
    \vspace{-10pt}
\end{figure*}

\begin{table*}
  \centering
\resizebox{\linewidth}{!}{
  \begin{tabular}{l|cccccc|cccccc|cccccc}
    \toprule
\multirow{2}{*}{Backbone} & \multicolumn{6}{c|}{RGB-Stacking-Strided} & \multicolumn{6}{c|}{Kinetics-Strided} & \multicolumn{6}{c}{DAVIS-Strided} \\
& $<\delta^{0}$ & $<\delta^{1}$ & $<\delta^{2}$ & $<\delta^{3}$ & $<\delta^{4}$ & $<\delta^{x}_{avg}$ & $<\delta^{0}$ & $<\delta^{1}$ & $<\delta^{2}$ & $<\delta^{3}$ & $<\delta^{4}$ & $<\delta^{x}_{avg}$ & $<\delta^{0}$ & $<\delta^{1}$ & $<\delta^{2}$ & $<\delta^{3}$ & $<\delta^{4}$ & $<\delta^{x}_{avg}$  \\
\midrule
\midrule
ResNet-18~\cite{he2016deep, doersch2023tapir}
& 36.6 & 64.6 & 82.7 & 90.9 & \textbf{94.7} & 73.9
& 10.5 & 35.3 & 65.7 & 81.3 & 88.6 & 56.3
& 9.7 & 31.1 & 60.6 & 78.3 & 87.0 & 53.3 \\

TSM-ResNet-18~\cite{lin2019tsm, doersch2022tap}
& 18.1 & 55.5 & 83.7 & 89.8 & 92.5 & 67.9
& 9.15 & 33.2 & 64.6 & 79.2 & 86.5 & 54.5
& 8.2 & 26.8 & 53.6 & 73.5 & 83.9 & 49.2 \\

CoTraker~\cite{karaev2023cotracker} Backbone
& 36.9 & 60.1 & 74.9 & 82.5 & 87.6 & 68.4
& 31.0 & 31.0 & 50.5 & 70.6 & 74.9 & 58.2
& \textbf{33.7} & 52.5 & 63.4 & 68.3 & 72.0 & 58.0 \\

DINOv2 (ViT-S/14)~\cite{oquab2023dinov2}
& 3.2 & 13.6 & 44.9 & 77.4 & 87.6 & 45.3
& 4.9 & 15.8 & 41.9 & 73.8 & 86.0 & 44.5
& 7.3 & 22.7 & 52.9 & 80.0 & 89.5 & 50.4 \\

DINOv2 (ViT-B/14)~\cite{oquab2023dinov2}
& 4.2 & 15.2 & 46.6 & 78.9 & 88.1 & 46.6
& 5.9 & 18.6 & 45.9 & 75.7 & 86.9 & 46.6
& 10.0 & 28.6 & 59.5 & 82.9 & 87.4 & 54.4 \\

\hlrow \textbf{\ours (ViT-S/14)}
& \underline{65.5} & \underline{81.7} & \underline{88.8} & \underline{92.3} & 93.7 & \underline{84.3}
& \underline{32.2} & \underline{55.2} & \underline{73.9} & \underline{84.0} & \underline{88.7} & \underline{66.8}
& 29.7 & \underline{56.4} & \underline{76.7} & \underline{86.8} & \underline{90.9} & \underline{68.0} \\

\hlrow \textbf{\ours (ViT-B/14)}
& \textbf{68.3} & \textbf{84.0} & \textbf{90.3} & \textbf{92.9} & \underline{94.5} & \textbf{86.0}
& \textbf{33.5} & \textbf{57.2} & \textbf{75.8} & \textbf{85.8} & \textbf{90.2} & \textbf{68.5}
& \underline{31.8} & \textbf{59.2} & \textbf{78.8} & \textbf{88.4} & \textbf{92.2} & \textbf{70.1} \\

\bottomrule
  \end{tabular}
}
\vspace{-5pt}
\caption{\textbf{Quantitative comparison on the TAP-Vid datasets~\cite{doersch2022tap} with the strided query mode.} Best scores are in bold and second best are underlined for each column.}
\label{tab:strided-quan}
\vspace{-15pt}
\end{table*}

\begin{table*}
  \centering
\resizebox{\linewidth}{!}{
  \begin{tabular}{l|cccccc|cccccc|cccccc}
    \toprule
\multirow{2}{*}{Backbone} & \multicolumn{6}{c|}{RGB-Stacking-First} & \multicolumn{6}{c|}{Kinetics-First} & \multicolumn{6}{c}{DAVIS-First} \\
& $<\delta^{0}$   & $<\delta^{1}$   & $<\delta^{2}$   & $<\delta^{3}$   & $<\delta^{4}$   & $<\delta^{x}_{avg}$ & $<\delta^{0}$   & $<\delta^{1}$   & $<\delta^{2}$   & $<\delta^{3}$   & $<\delta^{4}$   & $<\delta^{x}_{avg}$   & $<\delta^{0}$   & $<\delta^{1}$   & $<\delta^{2}$   & $<\delta^{3}$   & $<\delta^{4}$    & $<\delta^{x}_{avg}$\\
\midrule
\midrule
ResNet-18~\cite{he2016deep, doersch2023tapir} 
& 31.4 & 58.3 & 77.8 & 87.7 & \textbf{92.4} & 69.6 
& 8.6 & 28.8 & 56.5 & 74.2 & \underline{83.3} & 50.3 
& 9.0 & 27.3 & 54.9 & 73.7 & 84.1 & 49.8\\ 

TSM-ResNet-18~\cite{lin2019tsm, doersch2022tap} 
& 19.2 & 51.7 & 78.7 & 86.8 & 90.5 & 65.4 
& 7.8 & 28.1 & 55.2 & 71.4 & 80.2 & 48.6 
& 7.3 & 23.1 & 46.7 & 66.6 & 79.2 & 44.6\\ 

CoTraker~\cite{karaev2023cotracker} Backbone 
& 32.0 & 54.8 & 70.2 & 78.2 & 84.0 & 63.9
& 24.1 & 41.9 & 55.1 & 62.2 & 67.0 & 50.1
& \textbf{27.8} & 46.4 & 58.6 & 64.1 & 68.5 & 53.1 \\

DINOv2 (ViT-S/14)~\cite{oquab2023dinov2} 
& 3.0 & 11.8 & 38.9 & 72.2 & 84.1 & 42.0
& 4.2 & 13.6 & 36.3 & 65.9 & 79.2 & 39.8
& 6.0 & 19.9 & 47.5 & 73.7 & 84.5 & 46.3 \\

DINOv2 (ViT-B/14)~\cite{oquab2023dinov2} 
& 3.6 & 13.2 & 41.1 & 74.1 & 84.8 & 43.4
& 5.1 & 16.0 & 40.0 & 67.9 & 80.4 & 41.9
& 8.9 & 24.7 & 53.8 & 77.0 & 87.1 & 50.3 \\

\hlrow \textbf{\ours (ViT-S/14)} 
& \underline{58.2} & \underline{77.2} & \underline{85.4} & \underline{89.1} & 91.3 & \underline{80.2} 
& \underline{24.8} & \underline{46.2} & \underline{65.8} & \underline{77.5} & 83.1 & \underline{59.5} 
& 24.0 & \underline{49.2} & \underline{71.2} & \underline{82.8} & \underline{87.9} & \underline{63.0}\\

\hlrow \textbf{\ours (ViT-B/14)} 
& \textbf{61.8} & \textbf{79.6} & \textbf{86.9} & \textbf{90.2} & \underline{92.2} & \textbf{82.1}
& \textbf{26.0} & \textbf{48.4} & \textbf{68.2} & \textbf{79.8} & \textbf{85.3} & \textbf{61.6}
& \underline{26.1} & \textbf{52.6} & \textbf{74.5} & \textbf{84.9} & \textbf{90.0} & \textbf{65.6}\\

  \bottomrule
  \end{tabular}
}
\vspace{-5pt}
\caption{\textbf{Quantitative comparison on the TAP-Vid datasets~\cite{doersch2022tap} with the query-first mode.} Best scores are in bold and second best are underlined for each column.}
\label{tab:query-first}
\vspace{-10pt}
\end{table*}

\subsection{Implementation Details}
We implement our method using PyTorch~\cite{paszke2019pytorch}. During training, we use AdamW optimizer~\cite{loshchilov2017decoupled} with a learning rate of \(10^{-4}\), a weight decay of \(10^{-4}\), and a batch size of \(1\) per GPU. All models are trained for 100,000 iterations on 4 A100 GPUs, employing a cosine learning rate scheduler with warmup. For training, we use the Kubric Panning-MOVi-E dataset~\cite{doersch2023tapir,cho2024local} and sample \(256\) query points per batch. The hyperparameters are set as follows: the softmax temperature is \(\tau=20.0\), the soft argmax pixel threshold is \(M=5\) and the local temporal window size is \(N=13\), and the 2D convolution stride in the temporal adapter is \(s=4\). 

\subsection{Evaluation Protocol}
\label{sec: eval}

\paragraph{Evaluation datasets.} To assess our approach, we employ the TAP-Vid benchmark~\cite{doersch2022tap}, which comprises both real and synthetic video sets. The real videos come with accurate annotation tracks, while the synthetic videos are paired with perfect ground-truth trajectories. The benchmark consists of three datasets: TAP-Vid-RGB-Stacking, TAP-Vid-Kinetics, and TAP-Vid-DAVIS. TAP-Vid-RGB-Stacking~\cite{lee2021beyond} contains synthetic videos of a robot gripper stacking objects and includes 50 videos.
The TAP-Vid-Kinetics~\cite{kay2017kinetics} consists of 1,189 YouTube videos with diverse difficulties, including intense motion blur and abrupt scene transitions, while the TAP-Vid-DAVIS dataset~\cite{pont20172017} provides 30 videos featuring challenges such as substantial scale shifts of dynamic object.

\paragrapht{Evaluation metrics.}
We assess the accuracy of the predicted tracks using two metrics: position accuracy at various thresholds and average position accuracy \((<\delta^x_{\text{avg}})\). Position accuracy is evaluated at five threshold levels: \(<\delta^0\), \(<\delta^1\), \(<\delta^2\), \(<\delta^3\), and \(<\delta^4\), corresponding to accuracies within pixel distances of 1, 2, 4, 8, and 16, respectively. Each \(<\delta^x\) score represents the percentage of visible ground-truth points whose predicted positions fall within the specified threshold. The \(<\delta^x_{\text{avg}}\) score is the average across all thresholds.

Following~\cite{doersch2022tap}, we evaluate the datasets using two modes: strided query mode and first query mode. In strided query mode, query points are sampled along the ground-truth trajectory at intervals of 5 frames. In first query mode, query points are sampled from the first visible frame.
\subsection{Main Results}
\paragraph{Quantitative comparison.}
We evaluate our method against the backbones~\cite{he2016deep, lin2019tsm} commonly used in point tracking~\cite{cho2024local, doersch2022tap, doersch2023tapir, karaev2023cotracker, karaev2024cotracker3} and DINOv2~\cite{oquab2023dinov2} in both the strided query (Table~\ref{tab:strided-quan}) and the first query modes (Table~\ref{tab:query-first}).  Specifically, we compare with ResNet-18~\cite{he2016deep} from TAPIR~\cite{doersch2023tapir}, TSM-ResNet-18~\cite{lin2019tsm} from TAP-Net~\cite{doersch2022tap}, and the backbone from the pre-trained CoTracker model~\cite{karaev2023cotracker}, all of which are trained for point tracking. We report position accuracy to demonstrate the effectiveness of temporal information in our model.

Our small model, \ours (ViT-S/14) with DINOv2 (ViT-S/14), achieves state-of-the-art performance in position accuracy across most thresholds, outperforming other backbones on \(<\delta^x_{\text{avg}}\). \ours (ViT-B/14) with DINOv2 (ViT-B/14) delivers even better results. Across RGB-Stacking, Kinetics and DAVIS datasets, in the strided and first query modes, \ours (ViT-S/14) and \ours (ViT-B/14) consistently outperform competitors in PCK accuracy at most thresholds. The high accuracy at \(<\delta^0\) underscores the precision of \ours features in point tracking, with further gains at higher thresholds. At \(<\delta^4\), only \ours backbones achieve over 90\% accuracy on the DAVIS dataset in strided mode. Both \ours (ViT-S/14) and \ours (ViT-B/14) reach the highest position accuracy among backbones on average, demonstrating \ours’s ability to effectively model spatial and temporal information in videos.

\paragrapht{Qualitative comparison.}
We visualize the estimated tracks from the DAVIS dataset, with a qualitative comparison shown in Figure~\ref{fig:main_qual}. Unlike other methods, which produce highly jittery and inconsistent tracks over time, \ours generates temporally smooth and accurate tracks. This jitter in other methods is expected as they lack awareness of neighboring frames. In contrast, \ours utilizes temporal adapters to maintain frame-to-frame consistency, allowing it to produce smooth tracks even when individual frame estimates are less stable.

\subsection{Analysis and Ablation Study}
\label{sec: abl}

\paragraph{Comparison to two-stage point tracking models with iterative refiner.}
As shown in Table~\ref{tab:computation}, \ours, although solely a feature backbone without a refiner, achieves point estimation precision comparable to full pipelines with refiners, while demonstrating significantly higher efficiency in terms of throughput.
Unlike other models, which require refiners to inject temporal information for each query point, \ours embeds temporal awareness directly in the feature backbone. This design substantially boosts throughput: for example, \ours (ViT-B/14) achieves 12.5\(\times\) the speed of TAPIR with only a 3.5\%p drop in DAVIS accuracy, and even surpasses TAPIR in RGB-Stacking by 11.4\%p. Similarly, \ours (ViT-S/14) delivers 16.4\(\times\) higher throughput. These results demonstrate that \ours provides a simple yet highly effective solution, offering performance comparable to refiner-based models without requiring iterative processing. 
\begin{figure*}[h]
    \centering
    \includegraphics[width=\textwidth]{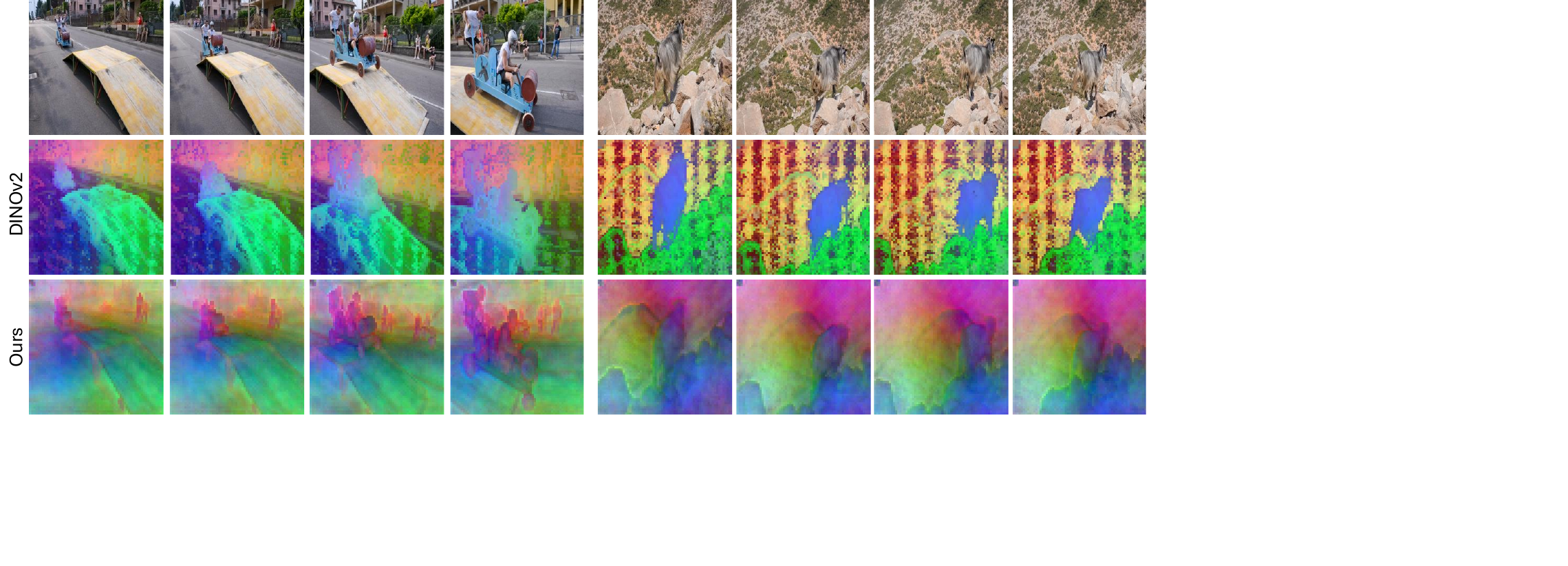}
    \vspace{-20pt}
    \caption{\textbf{Visualization of features.} We visualize the features generated by our model and DINOv2 using PCA. The results show that our model demonstrates improved temporal smoothness and finer-grained feature representation relative to DINOv2.}
    \vspace{-15pt}
    \label{fig:pca-vis}
\end{figure*}

\begin{table}
  \centering
\resizebox{\linewidth}{!}{
  \begin{tabular}{l|cc|ccc}
    \toprule
    \multirow{2}{*}{Method} & \multirow{2}{*}{RGB-Stacking-S} & \multirow{2}{*}{DAVIS-S} &   Throughput & \# of Refiner & Additional\\
     & & & (points/sec) & Params. & Params. \\
    \midrule
    \midrule
    RAFT~\cite{teed2020raft} & 58.6 & 46.3 &  23,405.71 & 4.2M & \underline{4.2M} \\
    TAP-Net~\cite{doersch2022tap} & 72.8 & 53.1 & \underline{29,535.98} & \underline{5.5K} & \textbf{5.5K}   \\ %
    TAPIR~\cite{doersch2023tapir} & 74.6 & \textbf{73.6} & 2,097.32 & 25.9M & 25.9M \\ %
    PIPs~\cite{harley2022particle}   & 51.0 & 59.4 & 46.43 & 26.0M & 26.0M \\ 
    \midrule
    \hlrow \textbf{\ours (ViT-S/14)} & \underline{84.3} & 68.0  & \textbf{34,396.30} & \textbf{0M} & 16.2M \\
    \hlrow \textbf{\ours (ViT-B/14)} & \textbf{86.0} & \underline{70.1}& 26,139.86 & \textbf{0M} & 26.0M \\

  \bottomrule
  \end{tabular}
  }
  \vspace{-5pt}
  \caption{\textbf{Comparison on point tracking pipelines using refiners.} \ours achieves competitive or superior accuracy while offering significantly higher throughput, despite not using a heavy iterative refiner. 
TAP-Vid-RGB-Stacking and TAP-Vid-DAVIS metrics report $<\delta^{x}_{avg}$. 
 Throughput is measured on a 24-frame video using a single NVIDIA RTX 3090 GPU.  
  }
  \label{tab:computation}
  \vspace{-10pt}
\end{table}

\begin{table}[t]
  \centering
  \resizebox{\linewidth}{!}{%
  \begin{tabular}{l|ccc|ccc|ccc}
    \toprule
\multirow{2}{*}{Method} & \multicolumn{3}{c|}{RGB-Stacking} & \multicolumn{3}{c|}{Kinetics} & \multicolumn{3}{c}{DAVIS}\\
&  AJ & $<\delta^{x}_{avg}$ & OA &  AJ & $<\delta^{x}_{avg}$ & OA &  AJ & $<\delta^{x}_{avg}$ & OA   \\
\midrule
\midrule
Kubric-VFS-Like~\cite{greff2022kubric}& 57.9 & 72.6 & 91.9 & 40.5 & 59.0 & 80.0 & 33.1 & 48.5 & 79.4 \\ %
TAP-Net~\cite{doersch2022tap}& 59.9 & 72.8 & 90.4& 46.6 & 60.9 & 85.0 &  38.4 & 53.1 & 82.3  \\ %
PIPs~\cite{harley2022particle}& 37.3 & 51.0 & {91.6}& 35.3 & 54.8 & 77.4 & 42.0 & 59.4 & 82.1 \\ %
RAFT~\cite{teed2020raft} & 44.0 & 58.6 & 90.4 & 34.5 & 52.5 & 79.7 &  30.0 & 46.3 & 79.6\\ %

TAPIR~\cite{doersch2023tapir}& 62.7 & 74.6 & 91.6 & 57.2 & 70.1 & 87.8 &  61.3 & 73.6 & 88.8 \\ %

FlowTrack~\cite{cho2024flowtrack}& - & - & - & - & - & - & 66.0 & 79.8 & 87.2  \\ %
CoTracker~\cite{karaev2023cotracker} & - & - & -& - & - & - &  65.9 & 79.4 & \underline{89.9} \\ %
LocoTrack~\cite{cho2024local}  & \underline{77.1} & \underline{86.9} & \underline{93.2}  & \underline{59.5} & \underline{73.0} & \underline{88.5}  & \underline{67.8} & \underline{79.6} & \underline{89.9} \\

\midrule
 \hlrow \textbf{\ours + LocoTrack} & \textbf{83.2} & \textbf{91.0} & \textbf{94.2} & \textbf{60.7} & \textbf{73.7} & \textbf{89.5} & \textbf{68.2} & \textbf{80.2} & \textbf{91.2}  \\
  \bottomrule
  \end{tabular}
  }
  \vspace{-5pt}
    \caption{\textbf{Quantitative comparison of \ours, adapted to LocoTrack~\cite{cho2024local}, on the TAP-Vid dataset with the strided query mode.} Our model shows a performance boost over LocoTrack on all datasets, with a particularly significant improvement on RGB-Stacking.
  }
  \label{tab:strided-quan-refiner}
  \vspace{-15pt}
\end{table}

\paragrapht{Analysis on \ours  as a backbone in existing point tracking pipeline.}
In Table~\ref{tab:strided-quan-refiner}, we present an analysis of the impact of adding a refiner after our model to investigate its ability as a feature backbone within a point tracking pipeline that uses iterative refinement~\cite{cho2024local,karaev2023cotracker,doersch2023tapir,harley2022particle}. To assess the ability to predict precise occlusion status, we use occlusion accuracy (OA) and the Average Jaccard (AJ) metric~\cite{doersch2022tap}, which measures both position and occlusion.

For our experiment, we incorporate the iterative refiner from LocoTrack~\cite{cho2024local}. Since LocoTrack utilizes three levels of hierarchical features from ResNet, we apply a simple convolutional upsampler to our single-resolution feature map. Specifically, using transposed convolution~\cite{zeiler2010deconvolutional}, we generate a 2$\times$ feature map with a 4$\times$ channel reduction and then a 4$\times$ feature map with a further 2$\times$ channel reduction. We freeze our model and train the convolutional upsampler and the LocoTrack refiner on the Kubric~\cite{greff2022kubric} panning MOVi-E dataset~\cite{doersch2023tapir}.

Our model combined with LocoTrack surpasses the original LocoTrack and also surpasses the performance of recent state-of-the-art point trackers~\cite{karaev2023cotracker,cho2024flowtrack}. Our model shows a boost on all datasets, and notably, it shows a huge boost in the RGB-Stacking dataset, achieving a +6.1 improvement in the AJ score. This result demonstrates the effectiveness of our approach when adopted into an existing point tracking pipeline.

\begin{table}[t]
    \centering
    \resizebox{\linewidth}{!}{
   \begin{tabular}{l|ccccccc}
        \toprule
        \multirow{2}{*}{Method} & \multicolumn{6}{c}{DAVIS}\\
        & $<\delta^{0}$ & $<\delta^{1}$ & $<\delta^{2}$ & $<\delta^{3}$ & $<\delta^{4}$ & $<\delta^{x}_{avg}$  
         \\
        \midrule\midrule
        1D Conv.  & 26.5 & 53.1 & 74.5 & 85.4 & \underline{90.1} & 65.9 \\
        3D Conv. & \underline{27.5} & \underline{54.3} & \underline{74.9} & \underline{85.5} & 90.0 & \underline{66.4} \\
        \midrule
        \hlrow \textbf{1D Attn. (Ours)} &\textbf{29.7} & \textbf{56.4} & \textbf{76.7} & \textbf{86.8} & \textbf{90.9} & \textbf{68.0} \\
        \bottomrule
\end{tabular}}%
\vspace{-5pt}
        \caption{\textbf{Ablation on temporal aggregation design choice.} 1D attention demonstrates the best performance compared to both 1D and 3D convolution layers.}
    \label{tab:temporal_agg}
    \vspace{-10pt}
\end{table}

\begin{table}[t]
    \centering
    \resizebox{\linewidth}{!}{
   \begin{tabular}{l|c|ccccccc}
        \toprule
        \multirow{2}{*}{Configuration} & \# of   & \multicolumn{6}{c}{DAVIS}\\
        & Adapters & $<\delta^{0}$ & $<\delta^{1}$ & $<\delta^{2}$ & $<\delta^{3}$ & $<\delta^{4}$ & $<\delta^{x}_{avg}$  
         \\
        \midrule\midrule
        Early Blocks  & 6 & 23.1 & 47.8 & 69.8 & 81.4 & 86.6 & 61.7 \\
        Later Blocks & 6 & 26.6 & 52.9 & \underline{74.3} & \underline{85.3} & \underline{90.1} & 65.8 \\
        Alternating Blocks & 6 & \underline{26.9} & \underline{53.4} & \underline{74.3} & \underline{85.3} & 90.0 & \underline{65.9} \\
        \midrule
        \hlrow \textbf{All Blocks} & 11 &\textbf{29.7} & \textbf{56.4} & \textbf{76.7} & \textbf{86.8} & \textbf{90.9} & \textbf{68.0} \\
        \bottomrule
\end{tabular}}%
\vspace{-5pt}
        \caption{\textbf{Ablation on the number and position of temporal adapter.} Rather than placing the temporal adapter into early or later blocks, or alternating blocks in DINOv2, applying it to all layers significantly boosts performance.}
    \label{tab:adapter_number}
    \vspace{-10pt}
\end{table}

\paragrapht{Ablation on temporal aggregation design choice.}
To model temporal information in our temporal adapter, we tested two approaches for temporal aggregation: convolution and attention. In Table~\ref{tab:temporal_agg}, we present an ablation study where only the temporal aggregation layer within \ours is modified, comparing 1D convolution, 3D convolution, and 1D attention, with all other components kept constant. While 1D convolution and 1D attention aggregate along the temporal axis only, 3D convolution considers both spatial and temporal dimensions. Due to computational constraints, this study is conducted on \ours (ViT-S/14).

The results suggest that 1D attention is effective for temporal aggregation, as it allows the model to capture motion dynamics by adapting to correlations between frames. This adaptability may contribute to improved tracking accuracy by enabling the model to weigh information from different frames and handle varied motion patterns and occlusions. In contrast, temporal convolutions, with their fixed weights, may be less effective in capturing these complex temporal relationships.

\paragrapht{Ablation on the number and position of temporal adapters.}
In Table~\ref{tab:adapter_number}, we conduct an ablation study on the placement of temporal adapters in \ours, which consists of 12 transformer blocks. We evaluate three configurations: placing six temporal adapters within the early six blocks, within the later six blocks, and in alternating blocks. 

Early transformer blocks often capture local details, while later blocks tend to focus on broader, global patterns and complex relationships~\cite{dosovitskiy2020image,raghu2021vision}. Therefore, it's plausible that adapters in the initial blocks may learn local motion cues, while those in later blocks might capture overall motion patterns. Placing adapters only in the early or later blocks might emphasize either local or global motion, respectively. While the alternating configuration shows some improvement, it may not be optimal. Integrating temporal adapters between each block could potentially allow for a more balanced capture of both local and global motion patterns, making features sensitive to multilevel motions and leading to better performance.
 
\begin{table}
  \centering
  \resizebox{\linewidth}{!}{
  \begin{tabular}{l|ccc}
    \toprule
    \multirow{2}{*}{Method} & \multirow{2}{*}{$<\delta^{x}_{avg}$} & Inference Time $\downarrow$ & GPU Memory $\downarrow$ \\
    &  & (sec) & (MiB) \\
    \midrule
    \midrule
    DINOv2 (ViT-S/14)~\cite{oquab2023dinov2} & 39.8 & \textbf{0.194} & \textbf{1,460} \\
    DINOv2 (ViT-B/14)~\cite{oquab2023dinov2} & 41.9 & \underline{0.443} & 2,816 \\
    \midrule
    \hlrow \textbf{\ours (ViT-S/14)} & \underline{68.0} & 0.575 & \underline{1,550} \\
    \hlrow \textbf{\ours (ViT-B/14)} & \textbf{70.1} & 1.197 & 2,998 \\

  \bottomrule
  \end{tabular}}
  \vspace{-5pt}
  \caption{\textbf{Computation and latency comparison with DINOv2.} We measure inference time and GPU memory usage for processing 24 frames without point prediction.}

  \label{tab:computation-per-image}
  \vspace{-15pt}
\end{table}

\paragrapht{Computation and latency comparison.} 
Table~\ref{tab:computation-per-image} compares our inference time and computation with that of DINOv2~\cite{oquab2023dinov2}, measuring the time and GPU memory taken to extract features for 24 frames. While our method exhibits comparable computation and slower inference due to temporal injection into the feature, \ours achieves a significantly higher score than vanilla DINOv2. This higher accuracy compensates for the increased time required for feature extraction.

\paragrapht{Visualization of feature with PCA.} We visually compare the feature of \ours and DINOv2~\cite{oquab2023dinov2} with PCA (Principal Component Analysis) in Figure~\ref{fig:pca-vis}. We reduce the feature dimensions to three, represented as RGB in the figure.

While DINOv2's visualizations demonstrate a noisy background and tend to show a uniform representation within a single object, our model produces consistent and smooth representations over time and also shows fine granularity within the same semantic object. We believe the temporally smooth representation and fine-grained feature detail lead to better point tracking.

\section{Conclusion}
In this paper, we presented \textbf{\ours}, a temporally-aware feature backbone for point tracking that integrates pre-trained DINOv2 representations with a temporal adapter, enabling long-term temporal context capture in feature space. Through extensive experiments, we demonstrate that embedding temporal information directly into the feature backbone, \ours reduces reliance on costly refiners while achieving both accuracy and efficiency. We anticipate \ours will inspire advancements in efficient, temporally-aware backbones for point tracking.

\paragrapht{Acknowledgements} This research was supported by Institute of Information \& communications Technology Planning \& Evaluation (IITP) grant funded by the Korea government (MSIT) (RS-2019-II190075, RS-2024-00509279, RS-2025-II212068, RS-2023-00227592) and the Culture, Sports, and Tourism R\&D Program through the Korea Creative Content Agency grant funded by the Ministry of Culture, Sports and Tourism (RS-2024-00345025, RS-2024-00333068), and National Research Foundation of Korea (RS-2024-00346597).
{
    \small
    \bibliographystyle{ieeenat_fullname}
    \bibliography{main}
}

\end{document}